\documentclass{article}

\PassOptionsToPackage{sort&compress}{natbib}
\usepackage[nonatbib, preprint]{neurips_2021}

\usepackage{color}

\usepackage[utf8]{inputenc}
\usepackage[T1]{fontenc}
\usepackage[colorlinks=true, citecolor=NavyBlue,linkcolor=OrangeRed,urlcolor=Rhodamine]{hyperref}
\usepackage{url}
\usepackage{booktabs}
\usepackage{amsfonts}
\usepackage{nicefrac}
\usepackage{microtype}
\usepackage{multirow}
\usepackage{algorithmic}
\usepackage{algorithm}
\usepackage{wrapfig}
\usepackage{graphicx}
\usepackage{comment}
\usepackage{amsmath,amssymb}

\usepackage{booktabs}
\usepackage{amsthm}

\usepackage{symbols}
\usepackage{multirow}
\usepackage{pmboxdraw}
\usepackage{mathtools}
\usepackage[table,xcdraw,dvipsnames]{xcolor}
\usepackage{thm-restate}

\usepackage{amsmath,amsfonts,bm}

\newcommand{\myparagraph}[1]{{\vspace{0.15cm}\noindent\bf #1}}

\def\1{\bm{1}}

\def\sp{space}

\def\vc{{\bm{c}}}

\def\ve{{\bm{e}}}
\def\vf{{\bm{f}}}

\def\vk{{\bm{k}}}

\def\vq{{\bm{q}}}
\def\vr{{\bm{r}}}

\def\vy{{\bm{y}}}

\def\mE{{\bm{E}}}
\def\mF{{\bm{F}}}

\def\mI{{\bm{I}}}

\def\mK{{\bm{K}}}

\def\mT{{\bm{T}}}

\def\mW{{\bm{W}}}
\def\mX{{\bm{X}}}

\DeclareMathAlphabet{\mathsfit}{\encodingdefault}{\sfdefault}{m}{sl}
\SetMathAlphabet{\mathsfit}{bold}{\encodingdefault}{\sfdefault}{bx}{n}

\def\gM{{\mathcal{M}}}

\def\gR{{\mathcal{R}}}

\def\gY{{\mathcal{Y}}}

\def\sR{{\mathbb{R}}}

\DeclareMathOperator*{\argmax}{arg\,max}

\definecolor{darkred}{rgb}{0.7,0.1,0.1}
\definecolor{darkgreen}{rgb}{0.1,0.7,0.1}
\definecolor{cyan}{rgb}{0.7,0.0,0.7}
\definecolor{dblue}{rgb}{0.2,0.2,0.8}
\definecolor{maroon}{rgb}{0.76,.13,.28}
\definecolor{burntorange}{rgb}{0.81,.33,0}
\definecolor{tealblue}{rgb}{0.212,0.459, 0.533}

\definecolor{pp}{rgb}{0.43921569, 0.18823529, 0.62745098}
\definecolor{rr}{rgb}{0.5254902 , 0.00784314, 0.12941176}
\definecolor{bb}{rgb}{0.09019608, 0.23529412, 0.37647059}
\definecolor{yy}{rgb}{0.49803922, 0.3372549 , 0.0}
\definecolor{gg}{rgb}{0.02352941, 0.3372549 , 0.17647059}

\usepackage{symbols}

\usepackage[multiple]{footmisc}
\usepackage{ctable}
\usepackage{pifont}

\definecolor{mybrown}{rgb}{0.87058824, 0.56078431, 0.01960784}
\definecolor{myblue}{rgb}{0.3372549 , 0.70588235, 0.91372549}
\definecolor{mypurple}{rgb}{0.8, 0.47058824, 0.7372549 }
\definecolor{myorange}{rgb}{0.835, 0.368, 0}
\definecolor{mygreen}{rgb}{0.00784314, 0.61960784, 0.45098039}
\definecolor{mygt}{rgb}{0.0078125 , 0.57421875, 0.40625}
\definecolor{mysp}{rgb}{0.84765625, 0.515625  , 0.0234375}

\title{Adapting CLIP For Phrase Localization Without Further Training}

\author{%
  Jiahao Li\;\;\;\;\;Greg Shakhnarovich\;\;\;\;\;Raymond A. Yeh\\
  Toyota Technological Institute at Chicago\\
  \texttt{\{jiahao, greg, yehr\}@ttic.edu} \\
}

\begin{document}

\maketitle

\begin{abstract}
Supervised or weakly supervised methods for phrase localization (textual grounding) either rely on human annotations or some other supervised models,~\eg, object detectors. Obtaining these annotations is labor-intensive and may be difficult to scale in practice. We propose to leverage recent advances in contrastive language-vision models, CLIP, pre-trained on image and caption pairs collected from the internet. In its original form, CLIP only outputs an image-level embedding without any spatial resolution. We adapt CLIP to generate high-resolution spatial feature maps. Importantly, we can extract feature maps from both ViT and ResNet CLIP model while maintaining the semantic properties of an image embedding. This provides a natural framework for phrase localization. Our method for phrase localization requires no human annotations or additional training. Extensive experiments show that our method outperforms existing no-training methods in zero-shot phrase localization, and in some cases, it even outperforms supervised methods. Code is available at \url{https://github.com/pals-ttic/adapting-CLIP}.

\end{abstract}

\section{Introduction}\label{sec:intro}
Phrase Localization (a.k.a. textual grounding) is the task of localizing bounding boxes referred by textual phrases in a given image. It has many down-stream applications,~\eg, visual question answering, image caption, and human computer interaction. 

With the advancement in deep learning models, supervised training of models has emerged as an dominant approach to build effective phrase localization systems~\cite{wang2016learning,HuCVPR2016,BryanPlummerICCV2017,wang2018learning,fukui2016multimodal,YehNIPS2017,plummer2018conditional,yang2019dynamic,yang2020improving,mu2021disentangled}. Critically, these methods require human annotations specific to textual grounding,~\ie, triplets of (image, phrases, bounding boxes). The annotation process to collect such a training set is expensive, labor intensive, and error-prone, especially, when a large dataset is necessary to effectively train these deep models. To address this, some efforts~\cite{rohrbach2016grounding,YehCVPR2018,wang2019phrase,wang2021improving,parcalabescu2020exploring} have considered weakly-supervised/unsupervised methods that rely on pre-trained supervised models,~\eg, object detectors, from other tasks~\cite{imagenet,everingham2010pascal,lin2014microsoft}.

Such pre-trained components are naturally biased towards the few commonly seen object categories, such as person, car,~\etc. These biases from the pre-trained model potentially lead to poor performance for uncommon or unseen objects in the training/pre-training set. To study these biases, ZSGNet~\cite{sadhu2019zero} propose zero-shot phrase localization, where the train and test sets have ``non-overlapping'' nouns hence evaluating models' zero-shot capability. In more details, they proposed different levels of ``non-overlapping'', at the word-level or at the category level. For example, ``Toyota'' and ``Honda'' would be non-overlapping at the word level, but considered overlapping at the category level. However, ZSGNet remains a supervised approach which requires human annotations on phrase localization.

To address the aforementioned issues, we propose a method that \textbf{does not rely on} any textual grounding, image classification or bounding box annotations. Instead, we leverage recent advances in large-scale contrastive language-vision model, CLIP~\cite{radford2021learning}, trained on a dataset of 400M image-text pairs collected from the internet. 
We propose a method to adapt/re-purpose CLIP for phrase localization. Importantly, our method can do so \textbf{without any extra supervision or training}, \ie, our method immediately improves with more advanced CLIP models. 

At a high-level, our method extracts high-resolution pixel-wise semantic features from images. By construction, these features lie in the same semantic space as the text embedding extracted using CLIP. As a result, we can compute per-pixel similarity scores with a given text query to obtain a heatmap. Localizing the described object in the text query becomes a score maximization problem,~\ie, finding a bounding box which achieves the highest score characterized by the heatmap. 

We demonstrate the effectiveness of our model on zero-shot phrase localization using Flickr30k Entities~\cite{plummer2015flickr30k} and Visual Genome (VG)~\cite{young2014image} datasets following the zero-shot setting proposed in ZSGNet~\cite{sadhu2019zero}. Our approach outperforms ZSGNet by an absolute $5\%$ on three out of four zero-shot splits over Flickr30k and VG, despite having never seen any textual grounding or object detection annotations. We also achieve comparable performance in long-tailed object categories compared to no-training methods that utilize pre-trained object detectors.

{\vspace{0.15cm}\noindent\bf Summary of our contributions:}
\vspace{-0.15cm}
\begin{itemize}
    \item We propose a method for textual grounding entirely from a pre-trained language-vision model (CLIP) without any bounding box supervision.
    \item We design methods for extracting high-resolution spatial feature maps from CLIP, for both ViT and ResNet architectures.
    \item We conduct extensive experiments and ablation studies demonstrating the effectiveness of our approach in zero-shot phrase localization.
\end{itemize}
\section{Related Works}
{\noindent \bf Phrase Localization.}
To study and evaluate the progress of phrase localization, numerous datasets, such as
Flickr30k Entities~\cite{plummer2015flickr30k}, Visual Genome~\cite{krishna2017visual} and  ReferItGame~\cite{KazemzadehEMNLP2014} have been proposed. These datasets contain a rich set of annotations covering a diverse set of objects and phrases.

Numerous approaches have been proposed to tackle the task of phrase localization~\cite{wang2016learning,HuCVPR2016,BryanPlummerICCV2017,wang2018learning,fukui2016multimodal,plummer2018conditional,yang2019dynamic,yang2020improving,mu2021disentangled}. These methods can be roughly divided into two groups: (a) Two-stage methods based on the the classical proposal-classification paradigm of object detection~\cite{BryanPlummerICCV2017,wang2017learning,hu2017modeling}. These take object proposals from the image and associated them with the corresponding query text, \eg, ranking these proposals based on embedding similarity to the text query. (b) In contrast, one-stage methods~\cite{endo2017attention, ChenKovvuriICCV2017} build an end-to-end pipeline without an intermediate proposal stage and are directly trained via bounding box regression.

Another line of work in phrase localization aims to reduce the requirement for textual grounding annotations~\cite{YehCVPR2018, wang2021improving} in a weakly supervised manner. A few efforts exist \cite{wang2019phrase,parcalabescu2020exploring} to build phrase localization models without any phrase localization data at all. Specifically, they utilized various off-the-shelf components such as detectors, word embedding models to create such systems. However, we note that these off-the-shelf detectors,~\etc, are themselves trained with human-annotated bounding boxes. 

\begin{figure*}[t]
    \centering
    \includegraphics[width=1.0\linewidth,]{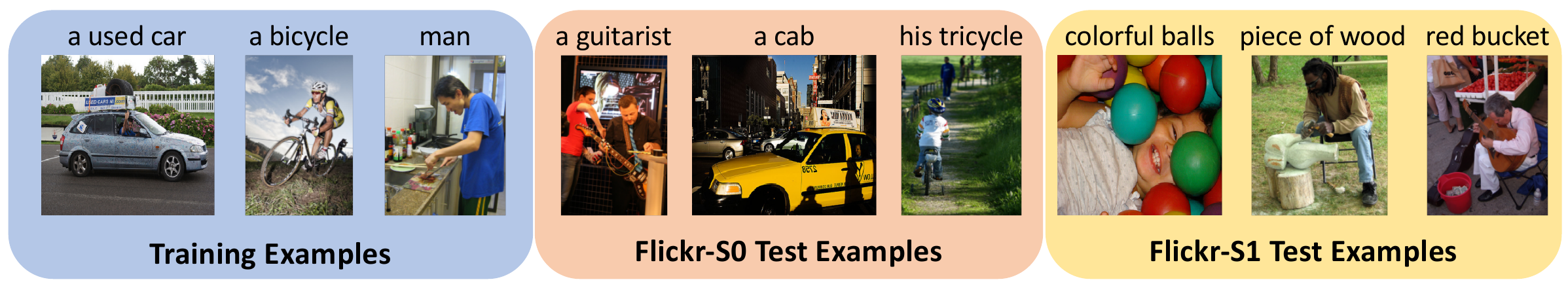}
   \vspace{-0.3cm}
    \caption{Illustration of different zero-shot phrase localization splits. ZSGNet~\cite{sadhu2019zero} proposes different level of semantic overlaps, specifically, in Flickr-S0 only the query phrase needs to be unseen~\eg, ``{\tt a used car}'' and ``{\tt a cab}'' can both refer to a car. However, this is not allowed in Flickr-S1, where the object category is also required to be unseen during training, ~\eg, ``{\tt red bucket}'' is not seen during training; both the category and the phrase is novel. 
    }
    \label{fig:zero-shot}
\end{figure*}

Finally, most relevant to our work is ZSGNet~\cite{sadhu2019zero} where they propose to evaluate the zero-shot generalization of phrase localization systems. Specifically, they re-split Flickr30K Entities and Visual Genome such that they have different levels of semantic overlap; see an illustrating example in~\figref{fig:zero-shot}. %
This helps quantify how well the model can generalize to completely unseen objects. The proposed ZSGNet remains a supervised model,~\ie, they still train on examples of (image, phase, bounding box). 
Different from ZSGNet, we propose to adapt a pre-trained CLIP for phrase localization. Our method does not require any annotations nor training (besides optional tuning of hyper-parameters on a validation set), while still achieving zero-shot phrase localization capacities. 

\myparagraph{Vision Language Models.}
Recently, large-scale contrastive models across language and vision domains have received a lot of attention,~\eg, CLIP~\cite{radford2021learning} and ALIGN~\cite{jia2021scaling}. Trained on $10^8-10^9$ of associated text and image pairs, these models aim to learn a joint embedding space between an image and natural language. These joint embeddings have demonstrated impressive capabilities of transferring to down-stream classification tasks without additional fine-tuning,~\ie, zero-shot classification. Hence, much of very recent work aims to use this capability beyond classification such as text conditioned image generation~\cite{liu2021more}, open-vocabulary and zero-shot detection~\cite{gu2021open,gu2021zero,zareian2021open} and segmentation~\cite{ghiasi2021open,zhou_arxiv2021_denseclip,xu2021simple}. %

Closely related to our work is DenseCLIP~\cite{zhou_arxiv2021_denseclip}, a concurrent work on ArXiv. They propose to extract spatial features from CLIP's ResNet image encoder. These features are then used to construct pseudo-labels for training another deep-net, to predict high-resolution semantic segmentation. While our method also generates spatial features from CLIP, our approach and goals differ: (a) we devise methods to extract spatial features from both the ResNet and (the better performing) ViT architectures, while DenseCLIP is limited to ResNet; (b) our method achieves high-resolution maps without the need of distillation or training; (c) we develop a framework for using the extracted spatial features for zero-shot phrase localization.

\section{Preliminaries: CLIP and Attention}\label{sec:prelim}
We provide a brief overview of CLIP and review Attention Pooling in detail to establish our notations. Familiarity with these concepts is necessary to understand our approach.

\myparagraph{CLIP.}
Recently, contrastive pre-training has emerged as a promising approach to learning effective image representations. CLIP proposes to train a contrastive model on 400 million image and text pairs~\cite{radford2021learning}. Specifically, given a paired image and text $\{(\mI_n, \mT_n)\}$, CLIP's objective is to find image embedding $\ve_{\tt img}(\mI_n) \in \sR^{D}$ and text embedding $\ve_{\tt txt}(\mT_n)\in \sR^{D}$, such that the cosine similarity between $\ve_{\tt img}$ and $\ve_{\tt txt}$ is maximized for a given input. With the embedding trained, CLIP demonstrates zero-shot transfer to downstream image classification tasks, typically by using the text prompt ``{\tt A photo of a \{label\}.}'', where ``{\tt \{label\}}'' is replaced with the category labels from the task. 

\myparagraph{Extracting Image Embedding with Attention Pooling.}
Given an image $\mI \in \sR^{H \times W \times 3}$, CLIP uses convolution layers to extract patch features $\mF \in \sR^{H'W' \times C}$, where the height and width is reduced to $H'\times W'$ and flattened into the first dimension. \Ie, $\vf_i \in \sR^{C}$ is a feature vector of patch $i$, commonly referred as a {\it patch token}. 

To create the image embedding, $\ve_{\tt img} \in \sR^{D}$, an additional {\it class token} $\vc$ is introduced. This class token is passed into attention layers along with the patch tokens $\mF$, and its corresponding output is the image embedding. For readability, we denote the class token as the $0^{\text{th}}$ patch token, \ie, $\vf_0 \triangleq \vc$. 
The computation for one attention layer is as follows:
\bea\label{eq:attn_pool}
\ve_{\tt img} = \sum_{i=0}^{H'W'} \alpha_i(\mW^Q\vc, \mW^K\mF) \cdot \mW^V \vf_i,
\eea
where $\mW^{Q/K/V}$ corresponds to a linear transformation to create the queries, keys and values of an attention layer. Next, $\alpha_i$ denotes the $i^{\text{th}}$ attention weight:
\bea
\alpha_i(\vq, \mK)  = \frac{\exp{(\vq^{\intercal}\vk_i/\tau)}  }{\sum_j \exp( \vq^{\intercal}\vk_j /\tau)},
\eea
where $\tau$ controls the temperature of the attention.
We point out that~\equref{eq:attn_pool} can be viewed as \textit{an operation over the set of patch tokens}. Specifically, shuffling the patch tokens' index results in the same $\ve_{\tt img}$; as along as the summation is over the same set.

CLIP's ResNet architecture uses multiple convolution layers followed by an attention layer. On the other hand, CLIP's ViT architecture uses a single convolution layer followed by several attention layers.

\begin{figure}[t]
    \centering
    \includegraphics[width=\linewidth]{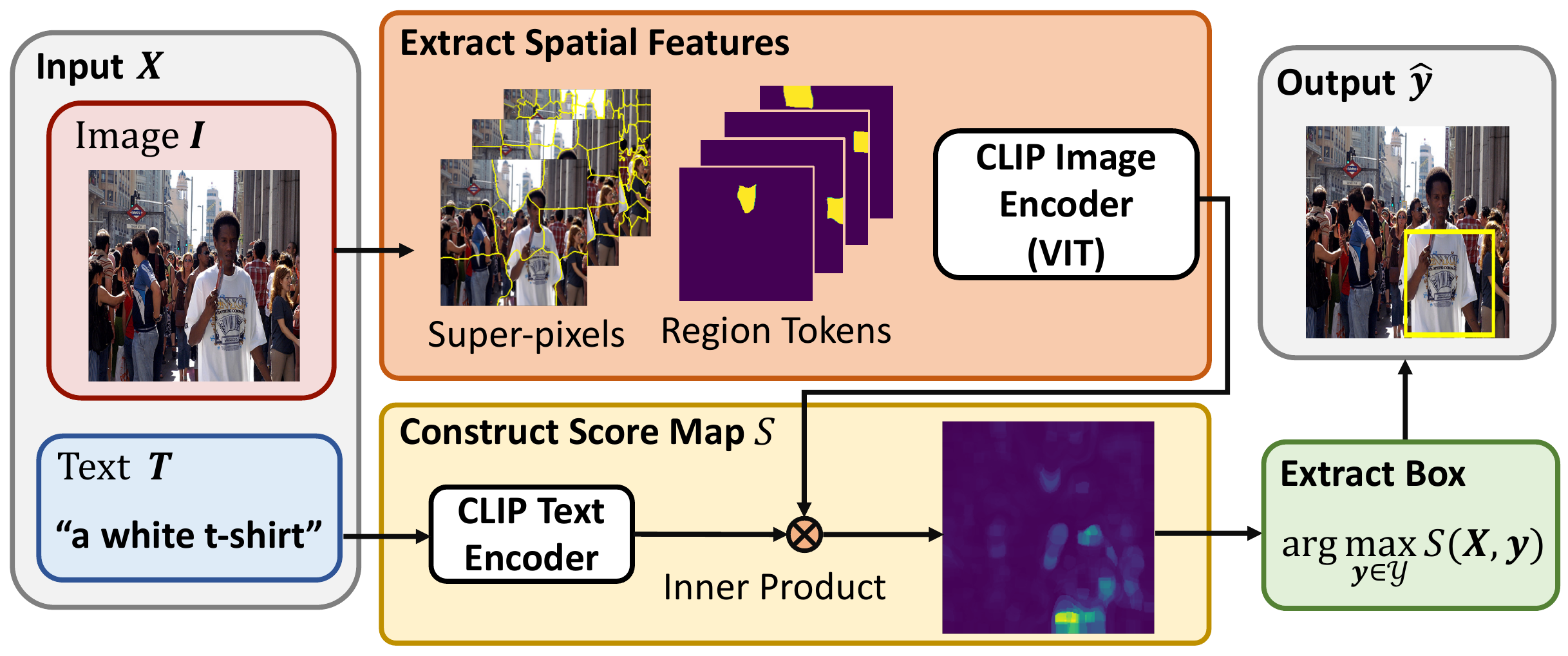}
    \vspace{-0.3cm}
    \caption{Proposed approach overview. Given a text query and image as input, we extract spatial features (per-pixel) and a text embedding using pre-trained CLIP. For each pixel location, we compute the cosine similarity between the spatial feature and text embedding resulting in a score map which we use to predict a bounding box formulated as a search over all possible bounding boxes.}
    \label{fig:overview}
\end{figure}

\section{Our Approach: Region-based Attention Pooling}\label{sec:app}
Our goal is to adapt a pre-trained CLIP model for the task of phrase localization. Recall, CLIP is trained to output an embedding vector for a given image or text phrase, hence the image embedding cannot be directly applied to phrase localization which requires spatial reasoning. 

To obtain spatial information, we propose to extract spatial features from CLIP for both ViT (\secref{sec:vit}) and ResNet (\secref{sec:resnet}) architectures. The key is to maintain the semantic meaning (\ie, alignment with language embeddings) of these spatial features to be the same as the original image embedding. After obtaining these spatial features, for each pixel location, we compute the inner product between the spatial feature and the text embedding extracted from CLIP to obtain a score map. Finally, we predict the bounding box that have the largest score according to the extracted map. An overview of our proposed method is depicted in~\figref{fig:overview}.

\subsection{Problem Formulation}
Given the input image and text query pair $\mX = (\mI, \mT)$, the task of phrase localization is to output the corresponding bounding box $\vy = (y_1,y_2,y_3,y_4)$, where a bounding box is defined by its top left $(y_1,y_2)$ and bottom-right $(y_3,y_4)$ corners. We formulate bounding box prediction as a score maximization problem:
\bea\label{eq:box_pred}
\hat\vy = \argmax_{\vy \in \gY} S(\mX, \vy),
\eea
where $\gY$ denotes the set of all possible bounding boxes. 

Our score function $S(\mX, \vy)$ is composed from a score map $\phi(\mX) \in \sR^{H \times W}$ extracted using CLIP. The score of a given bounding box $\vy$ is the aggregated values within $\vy$ on $\phi(\mX)$, \ie, 
\bea\label{eq:compute_score}
S(\mX, \vy) = \sum_{i=y_1}^{y_3}\sum_{j=y_2}^{y_4} \phi(\mX)_{ij} - R(\vy),
\eea
where $R(\vy) \triangleq \lambda \cdot \text{BoxArea}(\vy)$ is a penalty term regularizing the box size. 

In the remaining of this section, we will discuss how to effectively design this score map $\phi(\mX)$ from a pre-trained CLIP model without any additional training or data.

\subsection{Score Map Design}
Our score map $\phi(\mX)$ is constructed using features extracted from CLIP. Given an input $\mX = (\mI, \mT)$, we extract a text embedding from CLIP, $\ve_{\tt txt} \in R^{D}$. We also extract a ``spatial'', per-pixel, image embedding $\mE_{\tt img} \in \sR^{H \times W \times D}$ using CLIP (details deferred to~\secref{sec:vit} and~\secref{sec:resnet}). 

As in CLIP, we compute an inner-product between the text embedding and the per-pixel image embedding to relate text to image, \ie,
\bea\label{eq:norm_p}
\phi(\mX)_{ij} = \exp \left( \ve_{\tt txt}\cdot (\mE_{\tt img})_{ij} / \sigma \right).
\eea
where $\sigma$ denotes the temperature scaling parameter in CLIP. 

With the score map %
defined, we will next show how to extract a spatial per-pixel image embedding $\mE_{\tt img}$ from both CLIP's ViT and ResNet architecture. Recall, CLIP's image encoder outputs a vector of dimension $\sR^{D}$ without any spatial resolution.

\subsection{Spatial Embedding from ViT}
\label{sec:vit}
\paragraph{\bf Interpreting Attention Layers as Pooling.} As reviewed in~\secref{sec:prelim}, CLIP's ViT encoder consists of $L$ attention layers. At layer $l$, the class token embedding $\mathbf{e}^{(l+1)}$ is computed as
\begin{equation}
\label{eq:attn-pool-class}
\mathbf{\vc}^{(l+1)} = \sum_{i=0}^{H'W'} \alpha_i(\mW^Q\vc^{(l)}, \mW^K\mF^{(l)}) \cdot \mW^V \vf_i^{(l)},
\end{equation}
recall, for readability, the class token is also denoted as the $0^{\text{th}}$ patch token, \ie, $\vf_0^{(l)} \triangleq \vc^{(l)}$.

As can be seen, each attention layer outputs a weighted sum over all the patches, \ie, $i \in \{1 \hdots, H'W'\}$. In other words, the class token can be viewed as ``pooling'' information from all the patch tokens and itself. Hence, a forward pass of CLIP's ViT image encoder can be interpreted as an iterative pooling process as CLIP uses the final output $\mathbf{e}^{(L)}$ as the embedding vector for the entire image. 

\myparagraph{Modifying Attention for Spatial Features.} As our goal is to extract spatial image embedding, we need to perform pooling over \textit{image regions} instead of the entire image. For example, consider an image consists of a tree and a car. The attention layers would aggregate over both the tree and a car. This would lead to an embedding mixing the semantics of both objects, with spatial information lost. On the other hand, if we can just aggregate over the tree region and the car region separately, then the embedding would be spatially dependent. 

\begin{figure}[t]
    \centering
    \includegraphics[width=0.65\linewidth]{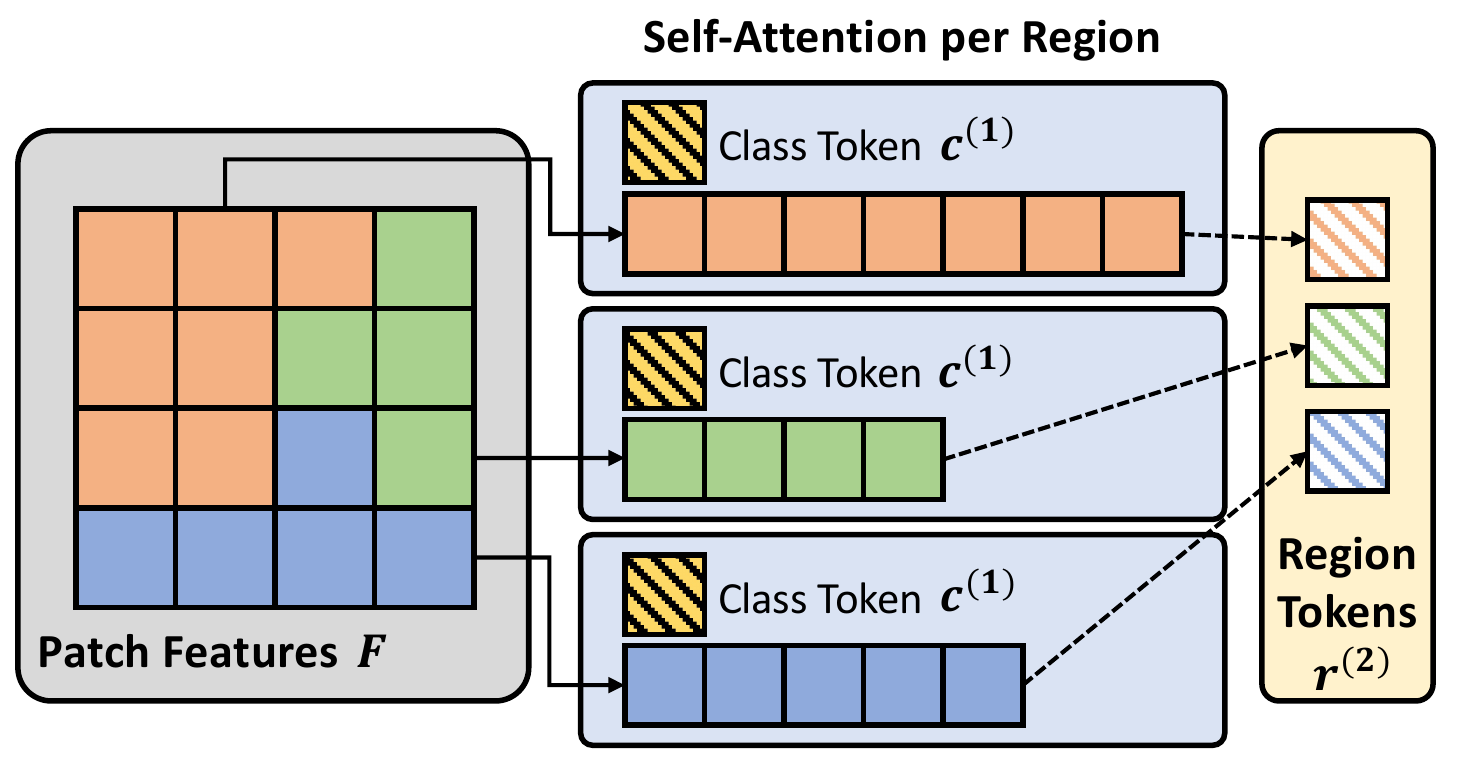}
    \caption{Illustration of the modified attention layer. Here, we illustrate the first attention layer in CLIP's ViT model. Specifically, we duplicate the class token $\vc^{(1)}$ as the initialization for region tokens $\vr^{(1)}$. Attention is perform over the patch features in each of the regions (color-coded) to produce an updated token $\vr^{(2)}$ for each region.
   }
    \label{fig:regions}
\end{figure}

The key question is how can we extract an embedding for a region, such that the embedding remains aligned with the text embedding $\ve_{\tt txt}$?
We propose to modify~\equref{eq:attn-pool-class} such that the class token only pools information from patches that are inside a region. %
Just as a class token $\vc$ which aggregates over the entire image, we introduce a \textbf{region token} $\mathbf{r}^{(l)}$ to aggregate over a region $\gR$. An input region token is initialized from the pre-trained class token, \ie, $\vr^{(1)} = \vc^{(1)}$. Different from class tokens, we only update $\vr^{(l)}$ by aggregating over the patches within $\gR$: 
\begin{equation}
\label{eq:attn-pool-region}
\mathbf{r}^{(l+1)} = \bigg(\sum_{i\in \gR} \alpha_i (\mW^Q\vr^{(l)}, \mW^V\mF^{(l)}) \cdot \mW^V \textbf{f}_i^{(l)}\bigg) + \alpha_\vr \cdot \mW^V \mathbf{r}^{(l)},
\end{equation}
where $\gR$ denotes a set of patch indices covered by the region. See illustration in~\figref{fig:regions}. Note that the patch tokens and class tokens are updated according to the standard CLIP, \ie, those are unaffected by the introduced region tokens. 
The operation in~\equref{eq:attn-pool-region} can be implemented as masked self-attention and computed efficiently in parallel.

To extract a $H\times W\times D$ feature map with per-pixel embedding vectors of size $D$.  First, we divide the image into a set of $\gM$ non-overlapping spatial regions $\gR_m$ using simple linear iterative clustering (SLIC)~\cite{achanta2012slic}, a classic super-pixel method. Next, we initialize a region token $\vr_m^{(1)}$ for each region in $\gM$. Following the aggregation in~\equref{eq:attn-pool-region}, we obtain the final embedding vectors $\vr_m^{(L)}, m=1, 2, ..., M$ for each region. 
Finally, to create a feature map of size $H\times W\times D$, pixel locations with a region is assigned the same final embedding vector, \ie,
\bea\label{eq:map_extract}
(\mE_{\tt img})_{ij} = \vr_m^{(L)} ~~\forall~(i,j) \in \gR_m.
\eea

This approach may be sensitive to number of super-pixels chosen in SLIC. To address this issue, we run SLIC at different ``resolution'',~\ie, using different numbers of super-pixels 
to extract multiple feature maps following~\equref{eq:map_extract} and then perform an pixel-wise average.

\myparagraph{Increasing Spatial Resolution.}
In CLIP's ViT architecture, the patch size is pretty large (32 or 16). As a result, the feature map is extracted at a substantially lower resolution $H^\prime\times W^\prime$. 
This can be problematic if the super-pixel covers a smaller area than a patch. %
To address this issue, we propose to modify the first convolution layer of ViT by reducing its stride. This can be viewed as extracting patch features with a sliding window, which leads to an increase in the number of patches,~\ie, increase in spatial resolution\footnote{The positional encoding of these new patches are bilinear interpolation of ones from CLIP.}. The remaining attention layers remain unchanged besides processing more patch features. %
As reviewed in~\secref{sec:prelim}, an attention layer is a ``set operation'' and the layer can process a larger set without any retraining.

\subsection{Spatial Embedding from ResNet}
\label{sec:resnet}

\paragraph{\bf Modifying Attention for Spatial Features.} CLIP's ResNet architecture consists of multiple convolution layers followed by a single attention layer for global pooling. An important difference between the two architectures is how the class token $\vc$ is trained. In ViT, the class token $\vc$ is randomly initialized and trained in an end-to-end manner. On the other hand, ResNet's class token is ``defined'' to be the average of all patch tokens, \ie, 
\bea
\vc \triangleq \frac{1}{H'W'} \sum_{i=1}^{H'W'} \vf_i.
\eea
Recall, for ResNet architecture, we refer to each vector (per spatial location) from the final convolution layer as a patch token.

Due to this difference, the construction of spatial features (per-patch) is more straight-forward in ResNet. Once again, in a standard attention layer, we can view a class token as aggregating information over all the patches. To obtain spatial features per patch, we need to aggregate over an individual patch. To do so, we introduce a ``class token'' for each patch $\vr_m^{(1)} \triangleq \vf_m$ (analogous to the region token described in~\secref{sec:vit}) and only aggregate over this patch:
\bea\label{eq:resnet_pool}
\vr_m^{(2)} = \sum_{i \in \{0,m\}} \alpha_i(\mW^Q\vr_m^{(1)}, \mW^K[\vf_0, \vf_m]) \cdot \mW^V \vf_i.
\eea
Recall the $0^{\text{th}}$ patch token corresponds to the ``class token'', \ie, for a single patch $\vf_0 \triangleq \vf_m$. Let's simplify~\equref{eq:resnet_pool} by substituting in $\vr_m^{(1)} = \vf_m$ and $\vf_0 = \vf_m$ to obtain (Note, $\{m, m\}$ denotes a set of duplicate elements):
\bea
\vr_m^{(2)} &=& \sum_{i \in \{m,m\}} \alpha_i(\mW^Q\vf_m, \mW^K[\vf_m, \vf_m]) \cdot \mW^V \vf_i\\
&=& 2 \cdot \alpha_m(\mW^Q\vf_m, \mW^K[\vf_m, \vf_m]) \cdot \mW^V \vf_m\\
&=& \mW^V\vf_m.\label{eq:resnet_pool_clean}
\eea
The last step uses the fact 
\bea
\alpha_m(\mW^Q\vf_m, \mW^K[\vf_m, \vf_m]) = \frac{\exp\left((\mW^Q\vf_m)^{\intercal}(\mW^K\vf_m)\right)}{\sum_{i \in \{m,m\}} \exp\left((\mW^Q\vf_i)^{\intercal}(\mW^K\vf_i)\right)} = 0.5.
\eea 
We note that this result in~\equref{eq:resnet_pool_clean} has also been discovered by a concurrent work, DenseCLIP~\cite{zhou_arxiv2021_denseclip}. Their discovery, however, is based on a hypothesis and empirical validation. Here, we present a mathematical justification of this result.
Finally, ~\equref{eq:resnet_pool_clean} enables the extraction of embedding vectors per patch, however, the spatial resolution is low due to large patch sizes. 

\myparagraph{Increasing Spatial Resolution.}
To address the low resolution issue when using~\equref{eq:resnet_pool_clean}, we aim to extract a higher resolution feature map without extra training, \eg, distillation in DenseCLIP~\cite{zhou_arxiv2021_denseclip}. 
The main cause of low-resolution feature map is the down-sampling/pooling layers in the Res-Net architecture which reduce the spatial resolution. 

Our method is inspired by Chen~\etal~\cite{chen2017deeplab}, where they remove the last few max pooling layers from a pre-trained  and use atrous (dilated) convolution to more densely sample the output. Following this idea, we replace all convolution layers in CLIP's image encoder with dilated convolution. Specifically, to get a high-resolution feature map, we increase the dilation factor by a multiple of two for every stride-two down-sampling/pooling in the original model. Pre-trained weights of the model is unmodified.

\subsection{Practical Considerations}\label{sec:practical}
In~\equref{eq:box_pred}, we formulated bounding box prediction as a search over all possible bounding boxes. Naively, using brute force search is infeasible. Existing branch and bound methods,~\eg, efficient sub-window search~\cite{LampertPAMI2009} is directly applicable. To further speedup this search, we devise a (greedy) hierarchical search strategy. This involves iteratively searching on a downsampled score map to find a coarse bounding box, then searching again by zooming in to this coarse box.
At each iteration, a brute-force search on all, low-resolution, bounding boxes based on integral images is computed using a GPU, which gives significant speedup. While this strategy does not guarantee the optimal solution, empirically we found it to perform well. 
\section{Experiments}\label{sec:exp}

\subsection{Experimental Setup and Details}
{\bf \noindent  Datasets.}
We follow ZSGNet's~\cite{sadhu2019zero} zero-shot setup on Flickr30k~\cite{BryanPlummerICCV2017} and Visual Genome~\cite{krishna2017visual} dataset for evaluation. Note that we only use these datasets for evaluation (and tuning two hyper-parameters) as our models do not require any training on phrase localization annotations. 
In more details, ZSGNet proposes four \textbf{zero-shot} splits based on Flickr30k and Visual Genome: 
\vspace{-0.15cm}
\begin{itemize}
    \item \textbf{Flickr-S0.} Phrases in the test set are not seen in the training set. But this split does not exclude the possibility that there are phrases in the training set describe objects in the same \textbf{object category}. %
    For example, if ``man" is in the training set then ``woman" is  allowed to be in the test set, even though both of these refer to a broader category of ``people''. As a result, this split is zero-shot for phrases, but \textbf{not for object categories.} This split corresponds to the Case 0 in ZSGNet~\cite{sadhu2019zero}.
    \item \textbf{Flickr-S1.} Phrases in the test set are not seen in the training set \textbf{and} there are no phrases in the training set belonging to the same object category of any text phrase. Flickr30k has several common object categories (\eg ``people", ``animals") and one ``other" category. ZSGNet uses all the examples in ``other" as validation and test sets, and examples of the remaining categories are used in the training set. This split corresponds to the Case 1 in ZSGNet~\cite{sadhu2019zero}.
    \item \textbf{VG-S0.} In this split of VG, phrases in the training and test sets are from different synsets~\cite{miller1995wordnet}. Also, no test images contain objects in the training synsets. This split corresponds to the Case 2 in ZSGNet~\cite{sadhu2019zero}.
    \item \textbf{VG-S1.} This is similar to VG-S0, except that each test image contains, in addition to the object to which the phrase refers, an object belonging to the training synsets. This split corresponds to Case 3 in ZSGNet~\cite{sadhu2019zero}.
\end{itemize}
\vspace{-0.15cm}
As not all prior works use these dataset splits, we also report results on standard, \ie, non zero-shot, split of the Flickr30k dataset:
\begin{itemize}
    \item \textbf{Flickr-All.} The original Flickr30k train-val-test split.
    \item \textbf{Flickr-Other.} The original Flickr30k subset split only containing objects belonging to the ``other" category. Note that this not the same as Flickr-S1 because ZSGNet reconstruct the splits over the entire train, validation, test splits. %
\end{itemize}

{\bf \noindent  Baselines.}
First, we compare against a straight-forward baseline based on CLIP, which we named, ``Crop \& Rank''. For an image-query pair, we first use a traditional bounding box proposal method (selective search~\cite{uijlings2013selective}) to get the top 200 bounding box proposals. For each proposal, we crop the corresponding image region and resize it to $224\times 224$. Next, we use CLIP to extract image embedding for each bounding box and the query's text embedding. The model predicts the bounding box where its cosine similarity with the text embedding is the highest out of all proposals.  

Next, we also compared with DenseCLIP~\cite{zhou_arxiv2021_denseclip} by using their method to extract spatial features. Different from our ResNet method, the resolution of the extract feature map is lower. The bounding box extraction procedure is identical to ours. Additionally, we also report performance on the following prior works:
\vspace{-0.15cm}
\begin{itemize}
    \item ZSGNet~\cite{sadhu2019zero} is a fully supervised method that reports on all of the standard and zero-shot splits.
    \item Wang~\etal~\cite{wang2019phrase} is a method that assembles off-the-shelf strongly supervised object detectors and word embedding models into a textual grounding model without extra training.
    \item Parcalabescu~\etal~\cite{parcalabescu2020exploring} is an improved model of Wang~\etal~\cite{wang2019phrase} by using additional structure information on image and text.
\end{itemize}
\vspace{-0.15cm}
For completeness, we report performance on 
fully supervised methods and methods using fully supervised pre-trained object detectors. However, this is not a fair comparison. Specifically, 
Wang~\etal~\cite{wang2019phrase} and Parcalabescu~\etal\cite{parcalabescu2020exploring} use supervised object detectors which naturally biased them to perform better on object categories that overlaps with the phrase localization dataset.

\myparagraph{Evaluation Metric.}
We adopt the widely used \textbf{Acc@\textit{thr}} metric for evaluation, where \textit{thr} refers to the threshold of intersection over union (IoU). A predicted bounding box is considered correct if its IoU with the ground-truth box is greater than the threshold. 
Acc@\textit{thr} reports the accuracy over the dataset,~\ie, the fraction of correct predictions over a dataset. 
We follow ZSGNet~\cite{sadhu2019zero} to use a threshold of 0.5 for Flickr30k and 0.3 for VG.

\begin{table}[t]
\centering
\caption{Quantitative Results on various splits of Flickr and Visual Genome datasets. Here, we report the Acc@\textit{thr} metric for each of the methods (the higher the better). The CLIP's architecture is in parentheses; RN50 corresponds to a ResNet architecture and ViT-L/14 corresponds to biggest Vit architecture released by CLIP.}
\label{tab:quan_results}
\setlength{\tabcolsep}{2pt}
\begin{tabular}{ cc | cc | c  c| cc  cc } 
\specialrule{.15em}{.05em}{.05em}
 \multirow{2}{*}{Method } & \multirow{2}{*}{Supervision } & \multicolumn{2}{c|}{Flickr} & \multicolumn{2}{c|}{Zero-Shot Flickr} & \multicolumn{2}{c}{Zero-Shot VG}\\
 & & All &  Other & S0 & S1 & S0 & S1\\
 \hline
 \hline
 ZSGNet~\cite{sadhu2019zero} & Full & \textbf{63.39} & \textbf{45.53} & \textbf{43.02} & 31.23 & 19.95 & 20.77 \\ 
 Wang~\etal~\cite{wang2019phrase} & BBox & 50.49 &  34.71 & - & - & - & -  \\ 
 Parcalabescu~\etal~\cite{parcalabescu2020exploring} & BBox & 57.08 &  38.52 & - & - & - & -  \\ 
  \hline
  Crop \& Rank (ViT-L/14) & CLIP & 4.21 &  6.34 & 7.17 & 7.51 & 17.24 & 16.97  \\ 
 DenseCLIP (RN50)~\cite{zhou_arxiv2021_denseclip} & CLIP & 34.26 &  25.87 & 31.08 & 27.78 & 14.71 & 15.68  \\ 
 \textbf{Ours (RN50)} & CLIP & 34.25 &  25.93 & 31.11 & 27.30 & 19.49 & 20.72  \\ 
 \textbf{Ours (ViT-L/14)} & CLIP & 43.80 &   35.42  & 40.37 & \textbf{36.10} &  \textbf{24.47} & \textbf{25.50} \\ 
\specialrule{.15em}{.05em}{.05em}
\end{tabular}
\end{table}

\myparagraph{Implementation Details.}
We use the latest-released CLIP model (ViT with patch size 14 and a $4\times$ scaled ResNet50) with our method\footnote{Pre-trained models are available at \url{https://github.com/openai/CLIP}}. We obtain regions by pooling SLIC maps with 100, 150, \ldots, 600 superpixels. %
The two parameters that have the biggest impact on the performance is the weighting factor $\lambda$ in the bounding box score~\equref{eq:compute_score} and the temperature term $\sigma$ in~\equref{eq:norm_p}. We tune these hyperparameters for different variants of our method on %
the validation sets of Flickr30k and VG respectively.

\begin{figure}[t]
\centering
\includegraphics[width=\linewidth]{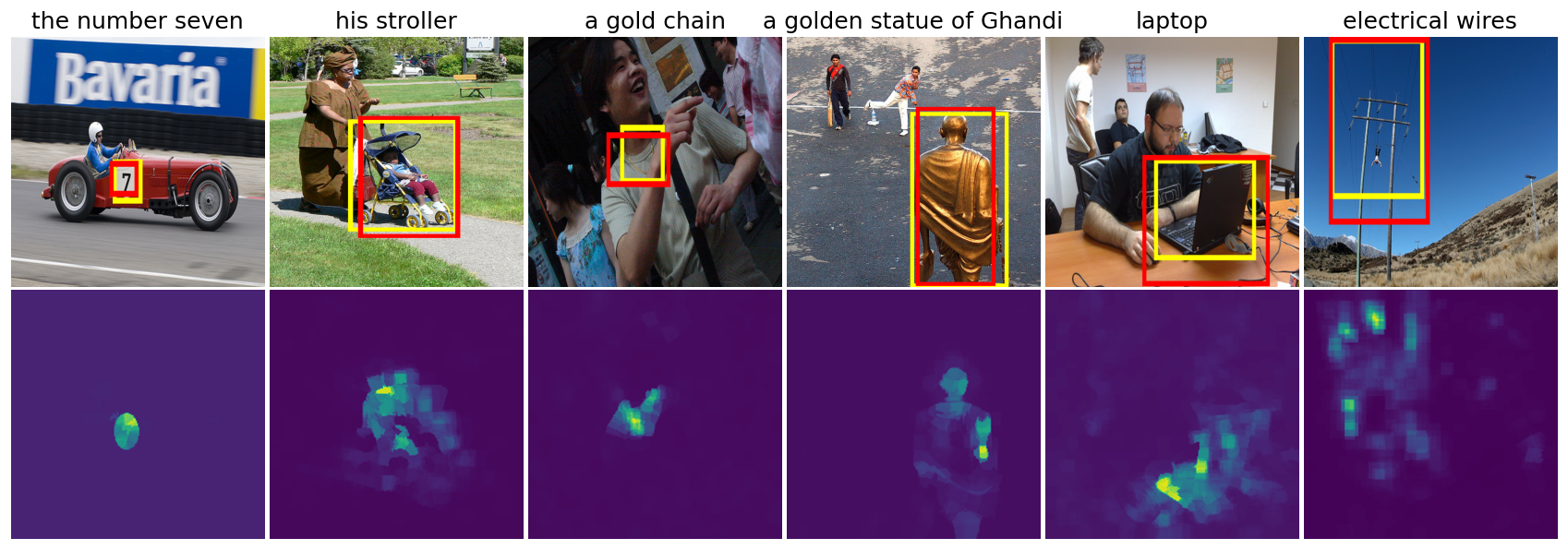}\\
\includegraphics[width=\linewidth]{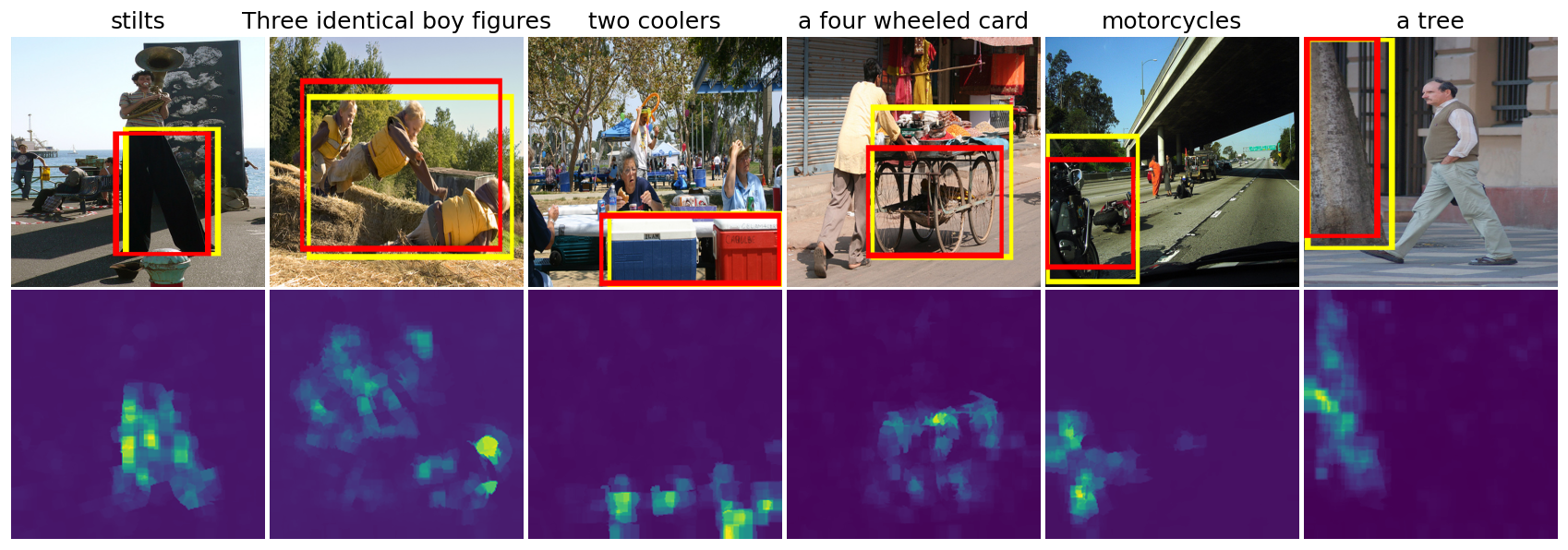}
\vspace{-0.3cm}
\caption{Qualitative results from our method. The query text is shown at the top. We visualize ground-truth bounding boxes in yellow and our predicted boxes in red. For each image, we also visualize the score map extracted from our method using CLIP's VIT-L14 architecture.}
\label{fig:success}
\end{figure}

\subsection{Quantitative Results}
We report quantitative comparisons with the baselines in~\tabref{tab:quan_results}. 

\myparagraph{Zero-Shot Results.}
Without any supervision from phrase localization datasets, our method
outperforms all compared baselines using CLIP supervision by a large margin. Additionally, our method even outperforms the fully supervised methods (ZSGNet~\cite{sadhu2019zero}) on Flickr-S1, VG-S1, VG-S2. We hypothesize that %
ZSGNet is negatively impacted by bias introduced by the training data distribution and have limited generalization capabilities. 

For Flickr-S0, our model achieves comparable performance to ZSGNet. We emphasize that Flickr-S0 is \textbf{not a truly zero-shot split}, because the same object category can appear in both the train and test set, while only the phrases are ``unseen''. This is more favorable to supervised methods because %
they can learn the visual features from the training data to better localize objects.

\myparagraph{Non Zero-Shot Results.}
For non-zero shot splits (Flickr-All and Flickr-Other), not surprisingly, supervised methods give the best performance. Our method achieves comparable performance to prior work that used supervised object detectors~\cite{wang2019phrase,parcalabescu2020exploring}.
Specifically, observe that the performance of ZSGNet on Flickr-All is much better than that of Flickr-Other. This again shows the bias introduced by the dominance of common object categories. On the other hand, the gap between ZSGNet and ours on Flickr-All is larger than on Flickr-Other. This shows that our method, which is not fine-tuned on any textual grounding dataset, is less prone to bias in object categories.

\subsection{Qualitative Results}
Beyond the quantitative results, we further study the quality of the extracted score map. In~\figref{fig:success}, we visualize some examples of our localization method on the Flickr30k dataset, including the score maps and the predicted bounding boxes. Ground-truth bounding boxes are shown in yellow and our predicted bounding boxes are shown in red. We observe that our method can correctly localize uncommon phrases such as ``stilts'' and ``electrical wires''.
This demonstrates our method's ability to handle all kinds of open vocabulary/world objects, including those rarely seen in fully-annotated object detection datasets.

\begin{figure}[t]
\centering
\includegraphics[width=\linewidth]{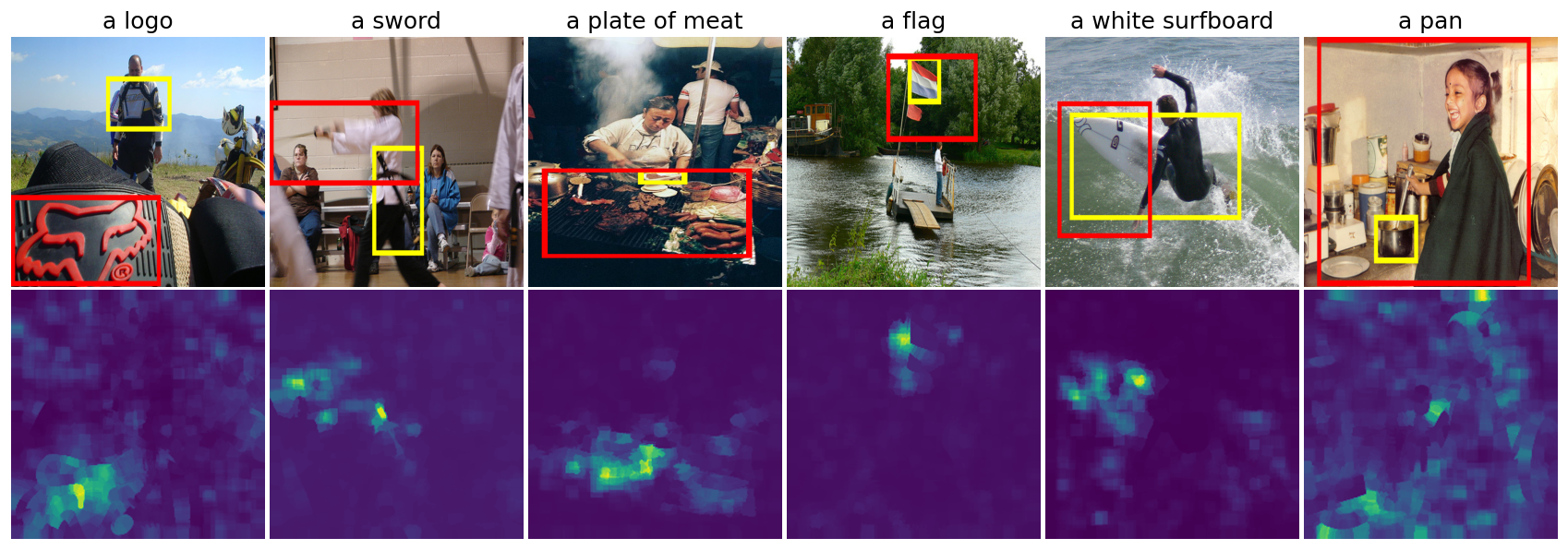}
\vspace{-0.4cm}
\caption{Failure cases of our method. Ground-truth in yellow and predicted box in red.}
\label{fig:failure}
\vspace{0.2cm}
\end{figure}

We also observe typical failure cases of our method, visualized  in~\figref{fig:failure}. %
First, annotations from the dataset (Flickr30k) may be ambiguous (see col. 1 and 2)
Second, our method tends to have challenges in localizing small objects (see col. 3 and 4). Specifically, for the image in column 4, the heatmap correctly identified the flag, however, the bounding box extraction procedure did not predict a small bounding box.
Third, we observe that our method may have challenges when handling occlusions (see col. 5), the score map only partially identified the surfboard. 
Finally, there are examples that the score map is not interpretable (see col. 6) and more investigation and understanding of CLIP embedding is necessary. 

\begin{table}[t]
\centering
\caption{Ablation study comparing different backbone architectures. We observe that the phrase localization performance improves as the CLIP model size increases. %
}
\label{tab:ablation_arch}
\setlength{\tabcolsep}{4pt}
\begin{tabular}{ ccc |  cc | cc  } 
\specialrule{.15em}{.05em}{.05em}
 \multirow{2}{*}{Architecture } & \multicolumn{2}{c|}{Flickr} & \multicolumn{2}{c|}{Zero-Shot Flickr} & \multicolumn{2}{c}{Zero-Shot VG}\\
 &  All &  Other & S0 & S1 & S0 & S1\\
 \hline
 \hline
 RN50  & 34.30 &  25.84 & 31.07 & 27.28 & 17.51 & 18.28  \\ 
 RN50$\times$4  & 34.25 &  25.93 & 31.11 & 27.30 & 19.49 & 20.72  \\ 
 ViT-B/32  & 29.49 &  24.72 & 27.48 & 25.00 & 13.43 & 14.21  \\ 
 ViT-B/16  & 41.54 &  35.29 & 37.82 & 35.70 & 18.23 & 19.19  \\ 
 ViT-L/14  & \textbf{43.80} &   \textbf{35.42}  & \textbf{40.37} & \textbf{36.10} &  \textbf{24.47} & \textbf{25.50} \\ 
\specialrule{.15em}{.05em}{.05em}
\end{tabular}
\end{table}

\subsection{Ablation Study}
We conduct ablation studies on two aspects of our method to show their effects on the phrase localization performance:
(a) how important is the quality of CLIP model; (b) how important is the spatial resolution of feature maps.

\myparagraph{CLIP Backbones.}
In~\tabref{tab:ablation_arch}, we show qualitative results for our method with different pre-trained CLIP architectures. It can be seen that larger models consistently outperform smaller models. This is most obvious for the ViT architecture. With smaller patch size, ViT patch tokens can capture more delicate spatial details of the scene, and therefore can better compute the features for some given region, even if it is small. Additionally, we observe that the best ViT CLIP model outperforms ResNet, validating our contribution over DenseCLIP~\cite{zhou_arxiv2021_denseclip} which is restricted to ResNet architectures.

\myparagraph{Feature Map Resolution.}
In~\tabref{tab:ablation_res}, we report an ablation study on the spatial resolution of feature maps. We report results with and without the upsampling technique for ViT. We consistently observe that upsampling to  higher resolution ($2\times$) feature maps results in better phrase localization performance. 

\begin{table}[t]
\centering
\caption{Ablation study on spatial resolution of feature maps. }
\label{tab:ablation_res}
\setlength{\tabcolsep}{4pt}
\begin{tabular}{ ccccc |  cccc   } 
\specialrule{.15em}{.05em}{.05em}
 \multirow{2}{*}{Architecture }
 &  \multicolumn{2}{c}{Flickr S0} &  \multicolumn{2}{c|}{Flickr S1} & \multicolumn{2}{c}{VG S0} & \multicolumn{2}{c}{VG S1} \\
 & $1\times$ & $2\times$ & $1\times$ & $2\times$ & $1\times$ & $2\times$ & $1\times$ & $2\times$\\
 \hline
 \hline
 ViT-B/32 & 27.05 & 27.48 & 23.80 &   25.00  & 12.79 & 13.43 &  13.46 & 14.21  \\ 
 ViT-B/16 & 34.10 &  37.82 & 29.26 &   35.70  & 13.51 & 18.23 &  13.87 & 19.19  \\ 
 ViT-L/14 & 34.67 &  40.37 & 30.30 &   36.10  & 18.92 & 24.47 &  19.54 & 25.50 \\ 
\specialrule{.15em}{.05em}{.05em}
\end{tabular}
\end{table}
\section{Conclusion}
We present a method for visual grounding by leveraging pre-trained language-vision model, CLIP, without any extra training. Our method generates high-resolution pixel-wise feature vectors from ViT and ResNet architectures of CLIP and computes per-pixel similarity scores with the embedding of the text query to create a score map. We then search over the score map to find the best box that localizes the object. Experiments on two datasets show that our method can effectively localize a wide variety of objects with complex text queries in the zero-shot setting. This method is a first step towards visual grounding based on large-scale pre-trained language-vision models without extra training, reducing potential biases caused by the limited size of human annotated datasets.

\bibliographystyle{ieeetr}
\bibliography{ref_spatial_clip}

\clearpage
\appendix
\renewcommand{\thetable}{A\arabic{table}}
\setcounter{table}{0}
\setcounter{figure}{0}
\renewcommand{\thetable}{A\arabic{table}}
\renewcommand\thefigure{A\arabic{figure}}

{\centering \Large \textbf{Appendix}}
\section{Additional Results}\label{sec:supp_qual}
\begin{figure}[h]
\centering
\includegraphics[width=\linewidth]{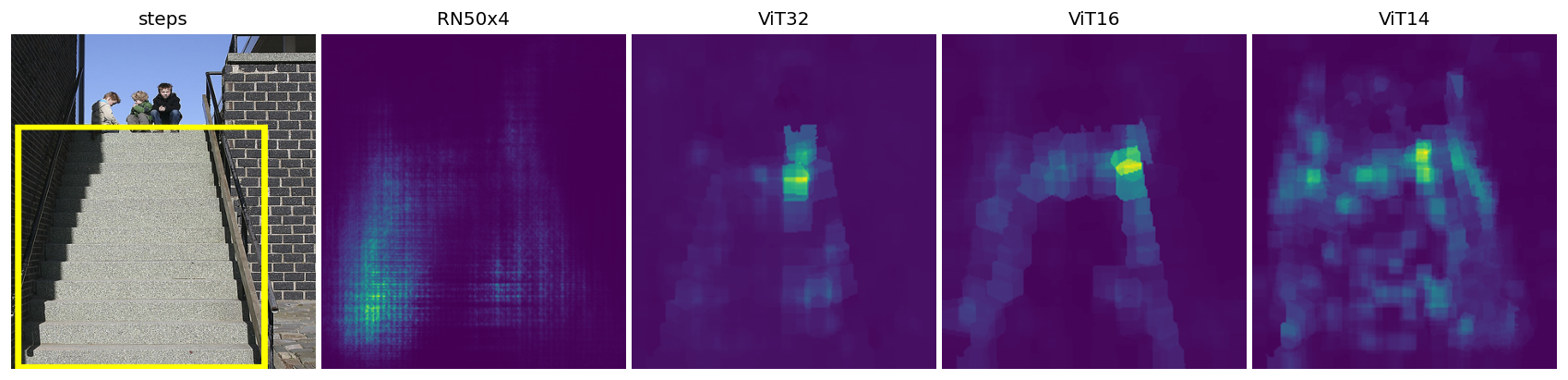}\\
\includegraphics[width=\linewidth]{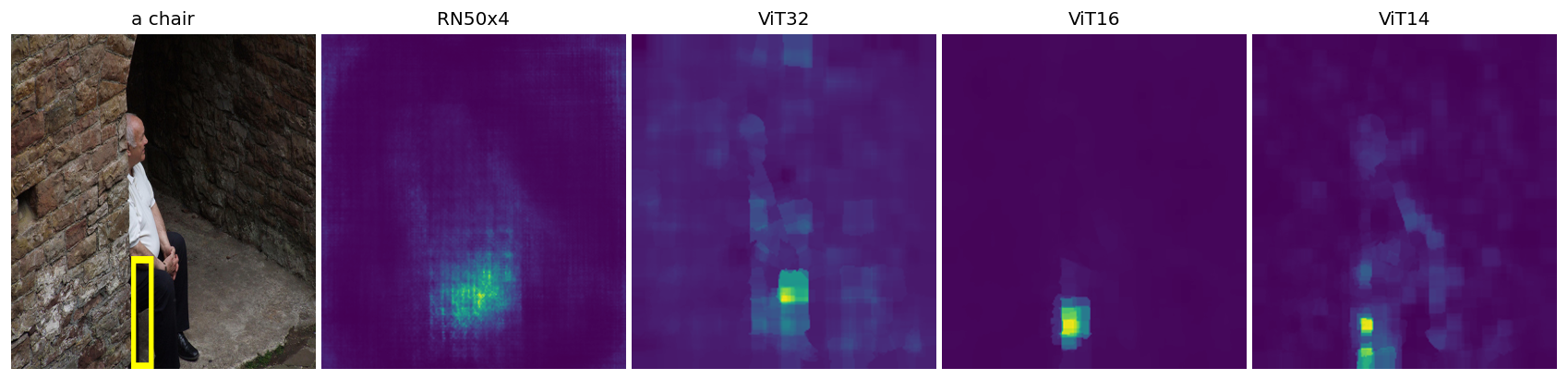}\\
\includegraphics[width=\linewidth]{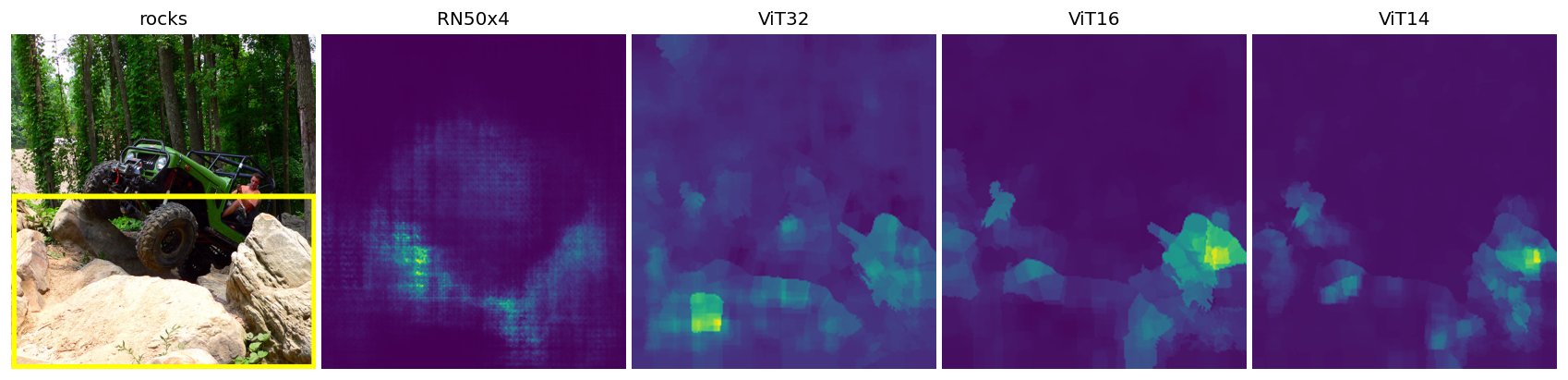}\\
\caption{Heatmaps extracted from different architectures}
\label{fig:arch_compare}
\end{figure}

{\bf\noindent Architecture comparison.}
In~\figref{fig:arch_compare}, we show comparisons of heatmaps extracted from different CLIP architectures. In general, we observe VIT architectures to perform better than the ResNet architecture, with ViT-L/14 performing the best. 

\begin{figure}[t]
\centering
\includegraphics[width=\linewidth]{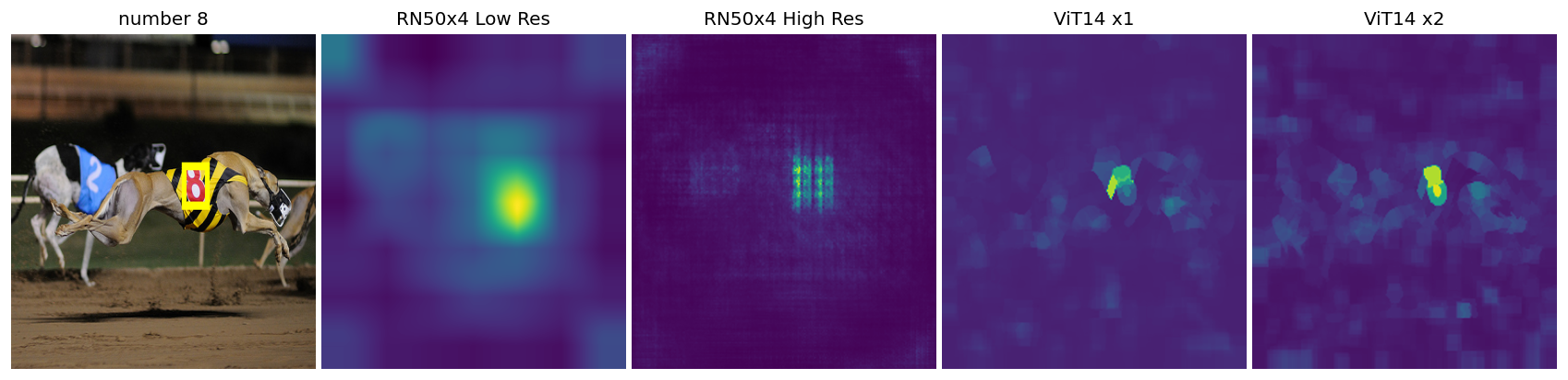}
\includegraphics[width=\linewidth]{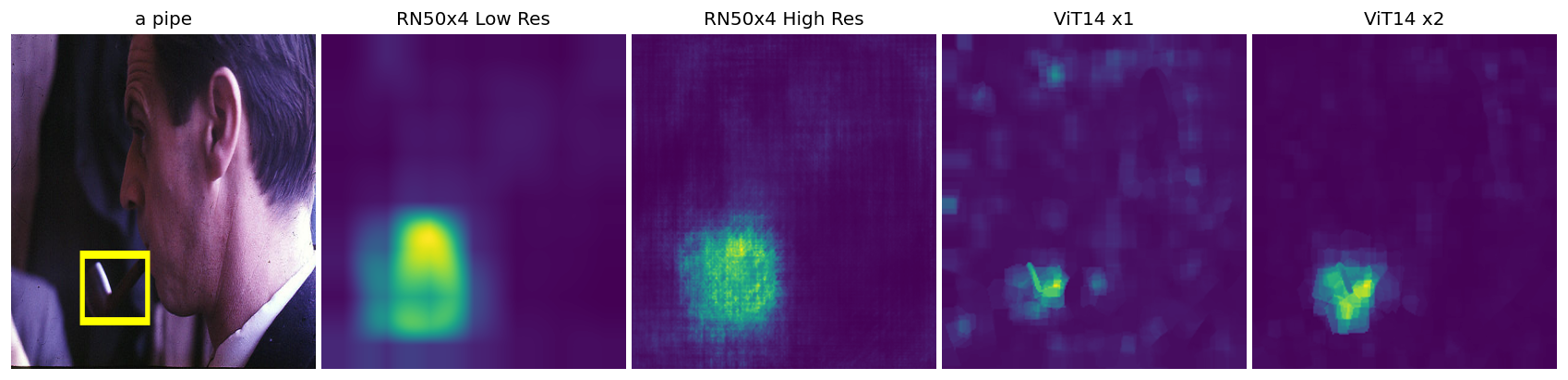}
\vspace{-0.3cm}
\caption{Heatmaps with different resolutions}
\label{fig:res_compare}
\end{figure}

\myparagraph{Resolution comparison.}
In~\figref{fig:res_compare}, we visualize our approach with and without using our technique for increasing the spatial resolution. For both ResNet and ViT architectures, we observe better localization of the object when using a higher resolution heatmap. 

\begin{figure}[h]
\centering
\includegraphics[width=0.9\linewidth, trim={0 0 7.25cm 0},clip]{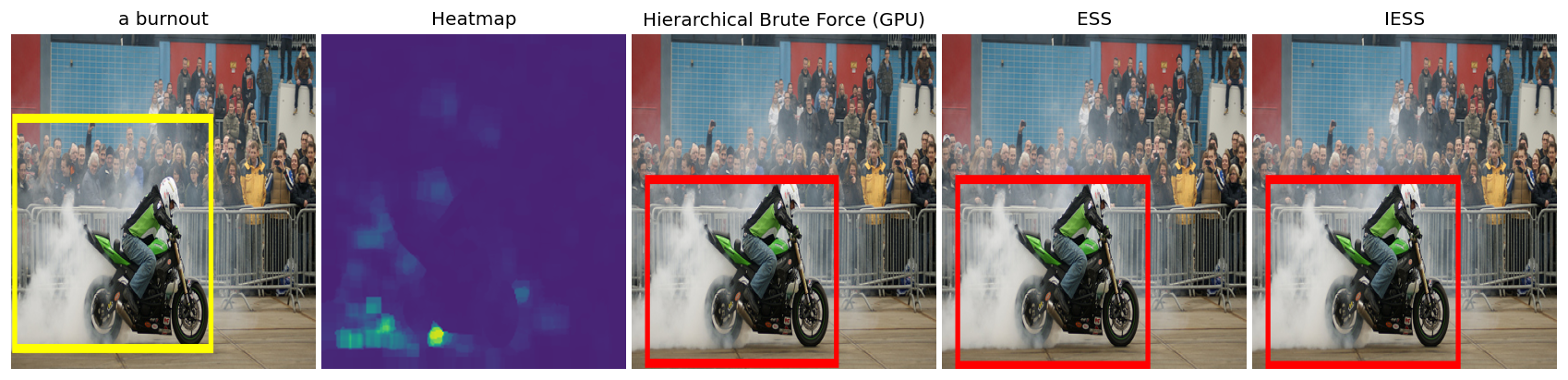}
\includegraphics[width=0.9\linewidth, trim={0 0 7.25cm 0},clip]{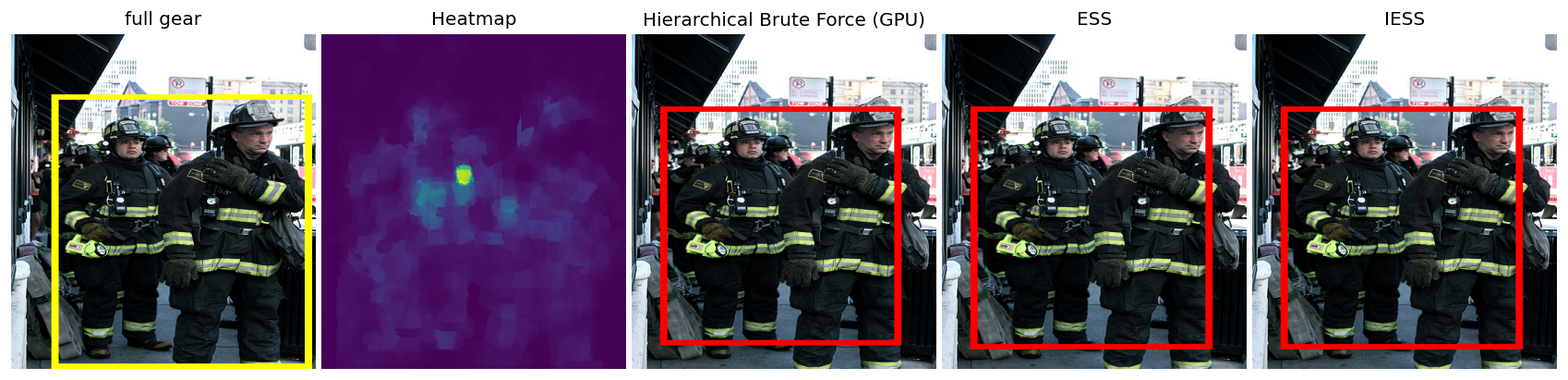}
\caption{Qualitative results with different box search methods}
\label{fig:bbox_compare}
\end{figure}

\myparagraph{Comparisons on bounding box prediction.}
In~\secref{sec:practical}, we describe a (greedy) hierarchical search strategy to find the best bounding box. While this greedy approach do not guarantee finding the best bounding box, empirically, we found it to perform similar to efficient sub-window search (ESS)~\cite{LampertPAMI2009}. In~\figref{fig:bbox_compare}, we observe that ESS and the hierarchical search results in nearly identical box predictions. In~\tabref{tab:ess}, we report the running time (mean and standard deviation over 100 examples) for the hierarchical search and ESS. We observe that the hierarchical search results in faster running time. Note the ESS implementation is in Python and runs only on CPU.

\begin{table}[h]
\centering
\caption{Comparison of bounding box search time}
\vspace{-0.1cm}
\label{tab:ess}
\setlength{\tabcolsep}{4pt}
\begin{tabular}{ cccc  } 
\specialrule{.15em}{.05em}{.05em}
 Method & Hardware & Mean Time (s) & Standard Deviation (s)\\
 \hline
 Hierarchical & RTX6000  & $2.3\times 10^{-3}$ &  $2.6\times 10^{-5}$   \\ 
 ESS (Python) & CPU & $3.1\times 10^{-1}$ &  $4.1\times 10^{-1}$   \\ 
\specialrule{.15em}{.05em}{.05em}
\end{tabular}
\end{table}

\begin{figure}[t]
\centering
\includegraphics[width=0.9\linewidth]{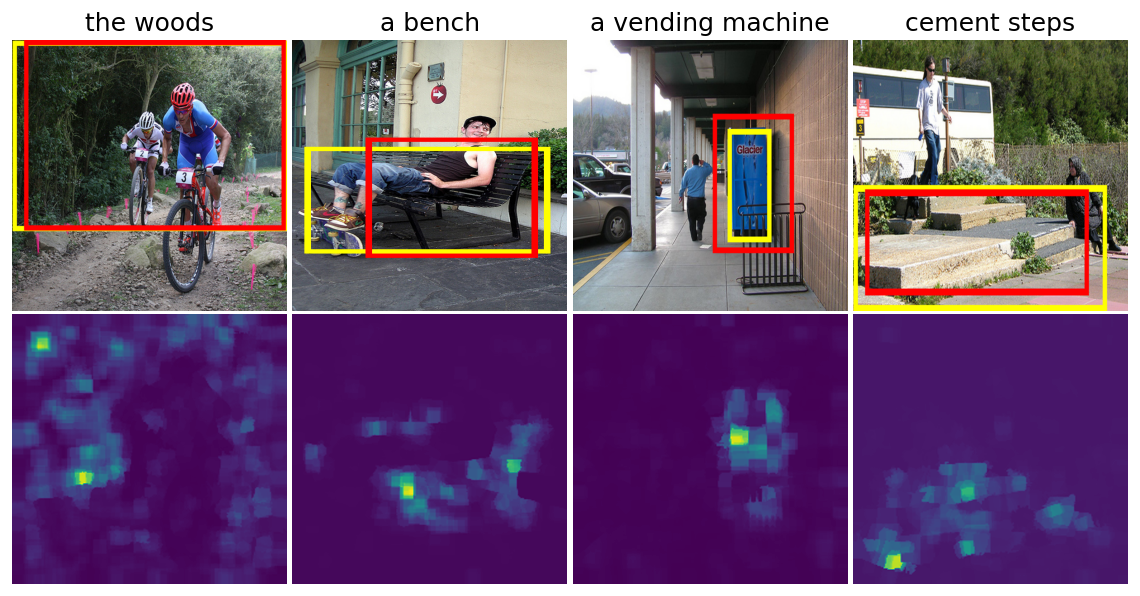}\\
\includegraphics[width=0.9\linewidth]{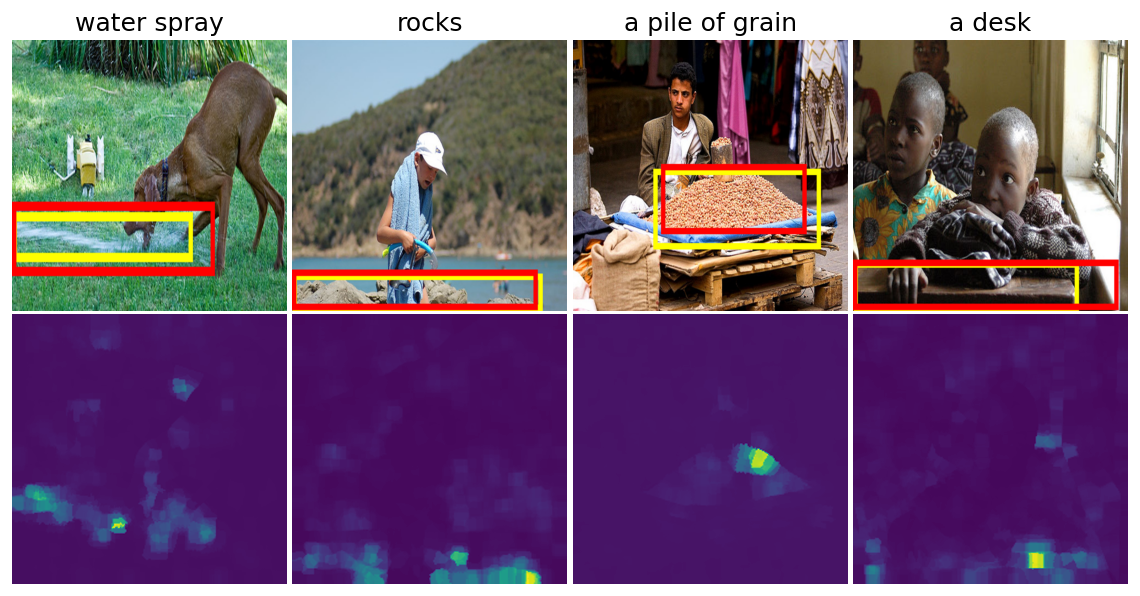}\\
\includegraphics[width=0.9\linewidth]{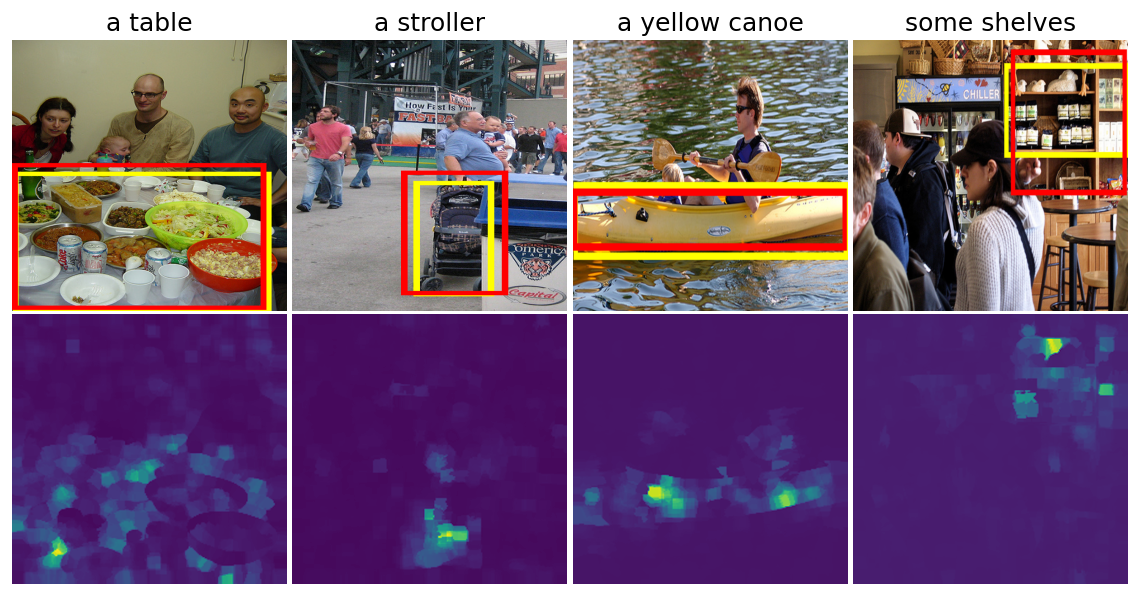}\\
\caption{Additional qualitative results.}
\label{fig:more_success1}
\end{figure}

\begin{figure}[t]
\centering
\includegraphics[width=0.9\linewidth]{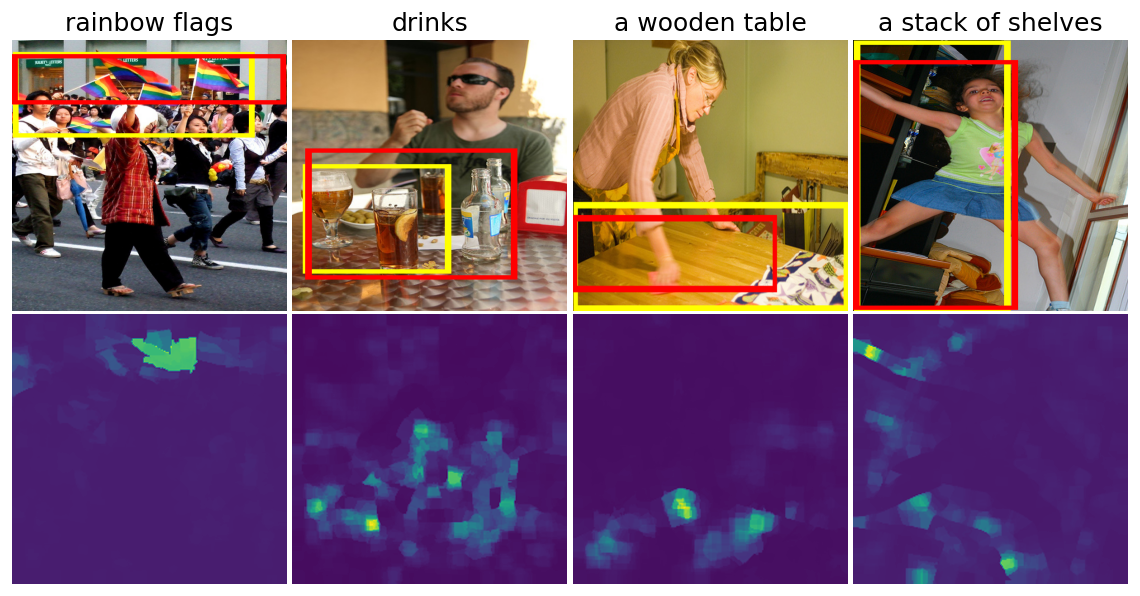}\\
\includegraphics[width=0.9\linewidth]{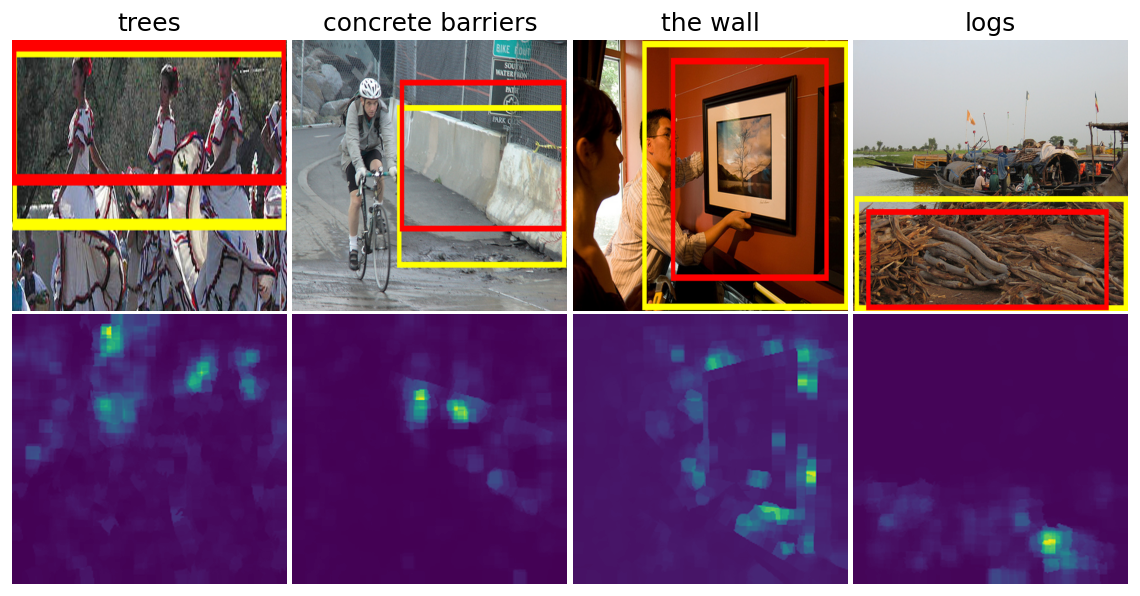}
\caption{Additional qualitative results.}
\label{fig:more_success2}
\end{figure}

\myparagraph{Additional Qualitative Result.}
We provide more qualitative results in~\figref{fig:more_success1} and ~\figref{fig:more_success2}. We observe that our approach successfully localizes a diverse set of phrases,~\eg, ``vending machine'', ``water spray'', ``rainbow flags'',~\etc. 

\end{document}